\documentclass[10pt,twocolumn,letterpaper]{article}

\usepackage[accsupp]{axessibility}  % Improves PDF readability for those with disabilities
\usepackage{iccv}              % To produce the CAMERA-READY version
\definecolor{iccvblue}{rgb}{0.21,0.49,0.74}
\usepackage[pagebackref,breaklinks,colorlinks,allcolors=iccvblue]{hyperref}
\usepackage{algorithm}
\usepackage{algorithmicx}
\usepackage{algpseudocode}
\usepackage{graphicx}
\usepackage{subcaption}
\algnewcommand{\LineComment}[1]{\State \(\triangleright\) #1}
\def\paperID{8} % *** Enter the Paper ID here
\def\confName{ICCV}
\def\confYear{2025}

\title{Zero-Shot Image Anomaly Detection Using Generative Foundation Models}

\author{Lemar Abdi, \:Amaan Valiuddin, \:Francisco Caetano, \:Christiaan Viviers, \:Fons van der Sommen \\
Eindhoven University of Technology
\\
The Netherlands\\
{\tt\small l.abdi@tue.nl}
}

\begin{document}
\maketitle
\begin{abstract}
% See if you like this:

Detecting out-of-distribution~(OOD) inputs is pivotal for deploying safe vision systems in open-world environments. We revisit diffusion models, not as generators, but as universal perceptual templates for OOD detection. 
% Then continue with your text
This research explores the use of score-based generative models as foundational tools for semantic anomaly detection across unseen datasets. Specifically, we leverage the denoising trajectories of Denoising Diffusion Models (DDMs) as a rich source of texture and semantic information. By analyzing Stein score errors, amplified through the Structural Similarity Index Metric (SSIM), we introduce a novel method for identifying anomalous samples without requiring retraining on each target dataset. Our approach improves over state-of-the-art and relies on training a single model on one dataset --- CelebA --- which we find to be an effective base distribution, even outperforming more commonly used datasets like ImageNet in several settings. Experimental results show near-perfect performance on some benchmarks, with notable headroom on others, highlighting both the strength and future potential of generative foundation models in anomaly detection.
\end{abstract}    
\section{Introduction}
\label{sec:intro}

Foundation models are large-scale general-purpose machine learning models trained on vast datasets to encapsulate broad, fundamental information applicable for multiple downstream tasks. One such downstream application is anomaly detection, which focuses on identifying data points that deviate from a preset in-distribution set (conventionally the training distribution). Often, the representational power of deep neural networks is utilized for this. Although traditionally framed as the identification of samples that deviate from the training distribution, anomaly detection has evolved with the rise of expressive models and large-scale datasets. This shift has moved the field away from task-specific feature engineering toward foundation models capable of detecting distributional shifts even between entirely unseen datasets. As a result, foundation models can be used as dataset-agnostic detectors to prevent overconfident predictions across a wide range of downstream tasks.

Since anomaly detection is largely unsupervised, generative models are commonly employed for this task. In particular, latent-variable models are a popular choice due to their ability to capture rich representations of the data~\cite{kingma2022autoencodingvariationalbayes}. Recently, Denoising Diffusion Models~(DDMs) have gained immense popularity due to their flexibility in general image generation. The strength of these architectures can be attributed to the finely discretized trajectory from latent space to data space, enabling the encapsulation of both a broad range of semantics as well as subtleties in the data~\cite{ho2020denoising, nichol2021improved, liu2022pseudo}. As such, the merits of this model design have been recognized in the field of anomaly detection~\cite{Serrà2020Input, mahmood2021multiscale, wolleb2022diffusion, graham2023denoising, liu2023unsupervised, mousakhan2024anomaly}. However, much of the existing literature remains narrowly focused on specific datasets, highlighting the need for further research into more generalizable, foundation model-based approaches.

In this work, we leverage DDMs as a generative foundation model~(GFM) for semantic anomaly detection by exploiting the statistics of the induced diffusion path. Specifically, we show that the data distributions can be distinguished by the error in the predicted Stein scores responsible for the diffusion trajectory. Additionally, we weigh the Stein scores by the Structural Similarity Index Measure (SSIM), emphasizing crucial, potentially anomalous areas that contribute to the score predictions. Our work can be considered an improvement to DiffPath~\cite{heng2024out}, and achieves state-of-the-art performance on various natural anomaly detection benchmarks. In contrast to conventional semantic anomaly detection, the proposed technique does not require any retraining or fine-tuning on the training set, truly leveraging the power of generative foundation models.
Our contributions can be summarized as follows.

\begin{itemize}
    \item We propose DiffPathV2, a method that that advances semantic anomaly detection using generative foundation models.
    \item We provide further insights on the Stein scores responsible for the induced diffusion trajectory.
    \item We achieve state-of-the-art performance on several natural image anomaly detection benchmarks.
\end{itemize}
% \vfill
The paper is structured as follows. We discuss related work in Section~\ref{sec:related} and theoretical background in Section~\ref{sec:back}. We propose our new methodology in Section~\ref{sec:method} followed by extensive experiments with ablations in Section~\ref{sec:results}. We finally conclude in Section~\ref{sec:conclusion}.

\section{Related Work}
\label{sec:related}

%% miccai paper related work? types of OOD detection

% Ja, OOD detection methods definieren en opsplitsen. Rough sketch:
\subsection{Unsupervised Image Anomaly Detection}

Out-of-Distribution~(OOD) Detection aims to identify inputs that deviate from the distribution $p_{\text{ID}}$ of a given in-distribution~(ID) dataset. Such problems are mostly unsupervised; no labels or examples of OOD data are available during training. This makes the task inherently challenging and highlights the importance of modeling the data distribution accurately. Existing OOD detection methods in vision can be broadly categorized into three methodological families:
\\
\par
% \noindent 
\textit{Distance-based methods} assume that OOD samples lie further from the ID learned embedding manifold. Hence, test samples are compared to in-distribution representations using metrics such as cosine similarity, Euclidean distance, or Mahalanobis distance. Recent work has improved robustness via ensemble representations or contrastive objectives~\cite{yang2024generalizedood}, although they often require storing feature banks or class prototypes, which makes scalability a concern. Furthermore, methods are often excessively sensitive and dependent on the learned feature space and especially unreliable in near-OOD settings~\cite{ren2021simple}.
\\
\par
\textit{Density-based methods} attempt to explicitly estimate the data log-likelihood using generative models. Approaches based on Normalizing Flows~(NFs), and Energy-based models~(EBMs) often fall in this category. Although intuitive and theoretically grounded, such methods often exhibit counterintuitive behavior, assigning higher likelihoods to OOD data~\cite{nalisnick_deep_2019, choi_waic_2019, kirichenko_why_2020}. To address these issues, recent work has proposed various alternatives. One direction involves using log-likelihood ratios~(LLR), which contrast the model likelihood under different hypotheses or reference distributions to mitigate the bias toward OOD regions. Another line of research reframes the detection task from evaluating the likelihood to the information-theoretic notion of `typicality'. In this view, samples are considered OOD if they are `atypical' with respect to the training distribution --- even if their likelihood is high. Typicality can be quantified using distance-based approximations~\cite{nalisnick_deep_2019, osada2024understanding} or via the Stein score of the log-likelihood of the data~\cite{chali2023improving, abdi2024typicality, viviers2025can}.
\\
\par
\textit{Reconstruction-based methods} can be considered as the implicit counterpart of density-based approaches. Instead of relying on likelihoods, generative models, such as a DDM or Variational Autoencoders~(VAEs), are evaluated on the reconstruction error subject to an OOD input signal. The reconstruction error is often quantified using the pixel-level MSE error, SSIM, or LPIPS. Image-level metrics can often be insufficient to quantify semantic data shifts. In fact, anomalous inputs have been shown to be reconstructed with high fidelity~\cite{zhou2022rethinking}, especially using natural imaging benchmarks~\cite{graham2023denoising}.

\subsection{Anomaly detection with DDMs}

Denoising diffusion models have recently gained traction for OOD detection, largely due to their strong generative fidelity and flexible, class-agnostic training paradigms. Initial diffusion-based approaches largely adhered to the reconstruction-based framework, where class-conditional models are trained to regenerate inlier examples~\cite{wolleb2022diffusion, mousakhan2024anomaly}. These reconstructions serve as baselines for error computation, either in pixel space or through learned representations. Other lines of work exploit the score-based nature of diffusion models to derive likelihoods through the ODE probability flow~\cite{song2021scorebased}. Although this enables exact likelihood computation in theory, it inherits the limitations similar to those of other likelihood-based methods, where high likelihoods can be assigned to outlier data. At the same time, such models often include high memory requirements.

The popularity of DDMs has introduced a novel perspective on anomaly detection. These methods analyze the \emph{forward} diffusion trajectory, that is, the noising process during inference, as an implicit representation of the underlying data manifold~\cite{heng2024out, sakai2025reconstruction}. The approach is based on the hypothesis that differing data marginals lead to significantly different intermediate noise trajectories across datasets. Among these methods, Heng~\etal~\cite{heng2024out} introduced \textsc{DiffPath}, a diffusion-based method capable of zero-shot OOD detection across natural image benchmarks without retraining or architectural changes. Their work demonstrates that the forward dynamics of a pretrained unconditional diffusion model, even when trained on unrelated data, encode sufficient statistics to discriminate between samples from different data distributions.

Our work --- \textsc{DiffPathV2} --- builds upon this by extending DiffPath, which further enhances discrimination performance across natural image benchmarks. These developments underscore a broader shift in the field: from task-specific anomaly detectors toward more general, reusable models that unify generative modeling and its downstream tasks with out-of-distribution detection in a single framework.

\section{Background}
\label{sec:back}

\begin{figure*}[t]
    \centering
    \includegraphics[width=\textwidth]{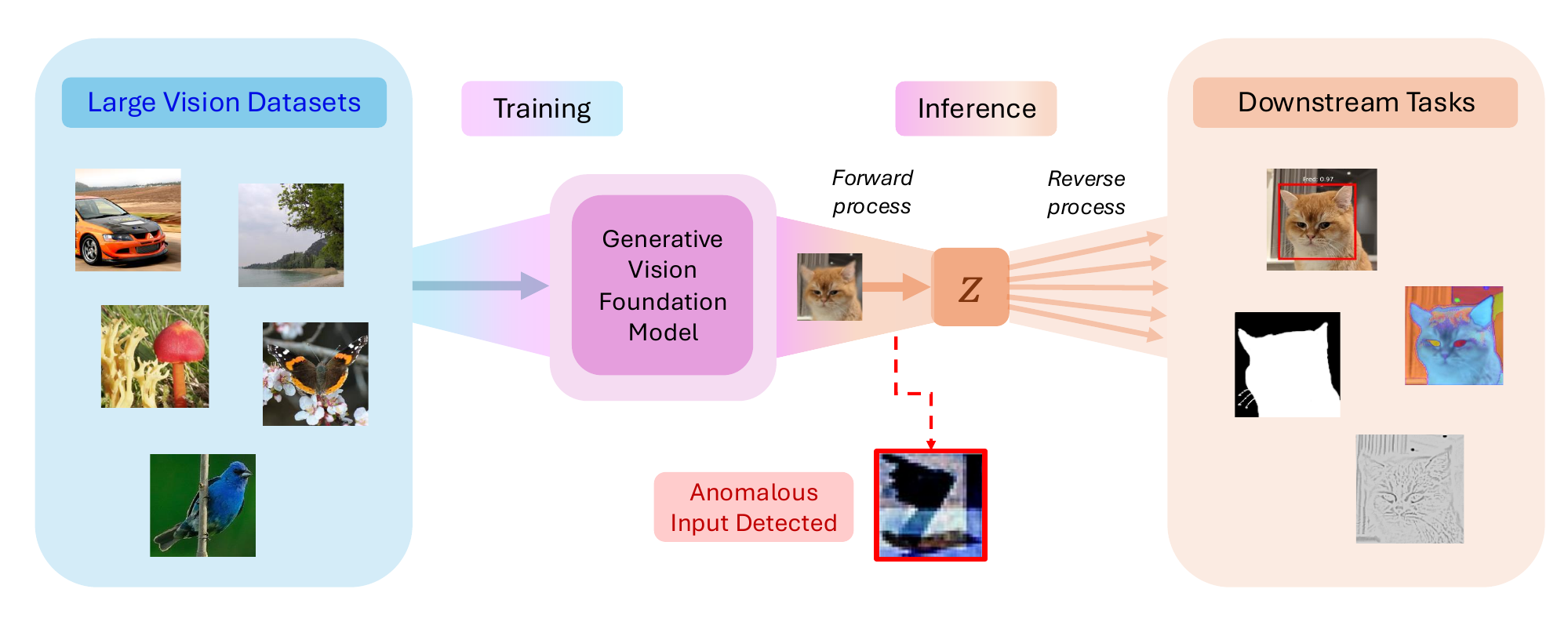}
    \caption{High-level overview of anomaly detection using generative foundation models}
    \label{fig:schem}
\end{figure*}

\subsection{Definitions}

% [hier waterproof maken wat wij definieren als anomaly detection]
% [reviewer kan moeilijk doen dus hier expliciet het volgende defineren zou handig zijn, volgens de framework van Yang et al, Generalized Out-of-Distribution Detection: A Survey]
 
To avoid any ambiguity and maintain terminological consistency, we adopt the taxonomy introduced by Yang~\etal~\cite{yang2024generalizedood}, which frames \textit{Generalized OOD Detection} as an umbrella task that includes three related subfields: \emph{anomaly detection}, \emph{open-set recognition}~(OSR), and \emph{outlier detection}. Anomaly detection~(AD) is defined as the task of detecting test-time samples that deviate from a predefined notion of normality characterized by the training data. 
 In \textit{sensory}~AD, the aim is to detect covariate shifts. Covariate shift induces abnormalities in high-level image statistics. AD problems focus on objects with similar or identical semantics, where varying imaging conditions or surface-level defects are sought to be identified. In contrast, \textit{semantic}~AD aims to detect test samples that exhibit entirely novel semantics, such as previously unseen object classes, which are not represented in the inlier distribution.

Given the generalized OOD detection framework, this work operates under the \emph{semantic anomaly detection} setting. Specifically, the unsupervised case is considered, where the model is trained only on unlabeled inlier data $\mathbf{x}$ sampled from $p_{\text{ID}}$. The task is to detect whether a test sample $\mathbf{x}^*$ originates from a semantically novel distribution $p_{\text{OOD}}$ that does not share the label space with the training data. Throughout this paper, the terms \textit{anomaly detection}, \textit{semantic anomaly detection}, and \textit{OOD detection} refer to this specific setting and are used interchangeably.

\subsection{Score-based diffusion models}

Score-based diffusion models gradually perturb the data $\mathbf{x}_0$ from the data distribution $p_0$ into a simple prior distribution $p_T$, typically a standard Gaussian, through a stochastic differential equation~(SDE). The forward diffusion process $\{\mathbf{x}_t\}_{t=0}^T$ is governed by the SDE
\begin{equation}\label{eq:sde}
\text{d}\mathbf{x} = \mathbf{f}(\mathbf{x}, t) \text{d}t + g(t)\text{d}\mathbf{w},
\end{equation}
where $\mathbf{f}(\mathbf{x}, t)$ denotes the drift term, $g(t)$ is a scalar diffusion coefficient, and $\mathbf{w}$ is a standard Wiener process.
Sampling from the learned data distribution involves reversing this diffusion process. The reverse-time SDE is given by
\begin{equation}
\text{d}\mathbf{x} = \left[ \mathbf{f}(\mathbf{x}, t) - g(t)^2 \nabla_\mathbf{x} \log p_t(\mathbf{x}) \right]\text{d}t + g(t)\text{d}\bar{\mathbf{w}},
\end{equation}
where $\nabla_\mathbf{x} \log p_t(\mathbf{x})$ is the \textit{Stein score} (or score function), and $\bar{\mathbf{w}}$ denotes the reverse Brownian motion over $[T, 0]$.
This stochastic process has been shown to have a deterministic counterpart, known as the probability flow ODE, which produces the same marginal distributions as the original SDE~\cite{song2021scorebased}. This formulation allows the diffusion process to be expressed as
\begin{equation}\label{eq:pf}
\text{d}\mathbf{x} = \left[ \mathbf{f}(\mathbf{x}, t) - \frac{1}{2}g(t)^2 \nabla_\mathbf{x} \log p_t(\mathbf{x}) \right]\text{d}t.
\end{equation}
Both the reverse-time SDE and the probability flow ODE require an accurate estimate of the score function. This is typically achieved by training a neural network $\boldsymbol{\epsilon}_\theta(\mathbf{x}, t)$ to approximate the Stein score function, using the score matching objective~\cite{vincent2011connection}
%
% \begin{equation}
% \min _\theta \mathbb{E}_{t \sim \mathcal{U}[0,T]}  \left\{ \mathbb{E}_{\mathbf{x}_0 \sim p_0(\mathbf{x}_0)} \mathbb{E}_{\mathbf{x}_t \sim p_{0t}(\mathbf{x}_t | \mathbf{x}_0)}\left[\left\|\boldsymbol{\epsilon}_\theta\left(\mathbf{x}_t, t\right)-\boldsymbol{\epsilon}\right\|_2^2\right] \right\}.
% \end{equation}
\begin{equation}
\min _\theta \mathbb{E}_{t}  \left\{ \mathbb{E}_{\mathbf{x}_0} \mathbb{E}_{\mathbf{x}_t}\left[\left\|\boldsymbol{\epsilon}_\theta\left(\mathbf{x}_t, t\right)-\boldsymbol{\epsilon}\right\|_2^2\right] \right\},
\end{equation}
where $t \sim \mathcal{U}[0,T]$, $\mathbf{x}_0 \sim p_0$, and $\mathbf{x}_t \sim p_{0t}(\mathbf{x}_t | \mathbf{x}_0)$.
The denoising target can be inferred by considering a linear Gaussian process at arbitrary time $t\sim\mathcal{U}[0,1]$.
\begin{equation}\label{eq:forwardnoise}
    \mathbf{x}_t = \sqrt{\bar{\alpha}} \,\mathbf{x}_0 + \sigma_t \, \boldsymbol{\epsilon}
\end{equation}
with source noise $\boldsymbol{\epsilon}\sim \mathcal{N}(0,1)$ and mixing parameter $\bar{\alpha}=\prod_s^t  \alpha_s$ and $\alpha_t=1-\beta_t$, with $\beta_t\in(0,1)$ being the variance schedule. Hence, the variance scheduling directly correlates with $\sqrt{\bar{\alpha}}$ and therefore the signal strength at timestep $t$. We can write forward noising process as $p_{0t}(\mathbf{x}_t | \mathbf{x}_0) = \mathcal{N}(\sqrt{\bar{\alpha}_t} \mathbf{x}_0, \sigma_t^2 \mathbf{I})$, take the derivative w.r.t. $\mathbf{x}_t$ to obtain
\begin{equation}
    \nabla_{\mathbf{x}_t}\log p(\mathbf{x}_t | \mathbf{x}_0) = - \frac{\mathbf{x}_t-\sqrt{\bar{\alpha}}\,\mathbf{x}_0}{\sigma_t},
\end{equation}
with $\sigma_t^2 = (1 - \bar{\alpha}_{t-1})/(1 - \bar{\alpha}_t)$, following the DDPM formulation~\cite{ho2020denoising} for the fixed time-dependent variances. Substitution of Equation~\ref{eq:forwardnoise} followed by multiplication with $\sigma_t$ results in the ground-truth score
\begin{equation}
    \boldsymbol{\epsilon} = -\sigma_t \nabla_{\mathbf{x}_t} \log p_{0t}(\mathbf{x}_t | \mathbf{x}_0).
\end{equation}
\section{Methods}
\label{sec:method}

As mentioned in Section~\ref{sec:intro}, our framework takes inspiration from {DiffPath}~\cite{heng2024out} and is built upon the assumption that different data distributions will exhibit distinctive denoising dynamics along the diffusion trajectory. The established theoretical framework of score-based DDMs in the preceding section enables us to precisely define the quantification of the trajectory dynamics. 

\subsection{DiffPathV2}
We train a time-conditional encoder-decoder denoising model, $\boldsymbol{\epsilon}_\theta(\mathbf{x}_t, t)$, on a large-scale data set to predict scores at each timestep. In previous work, DiffPath extracted these predicted scores to quantify statistics in the denoising trajectory. However, we argue that it is more principled to model the estimated score \textit{error}, rather than just the scores themselves.  If a test sample $\mathbf{x}_0$ is drawn from $p_{\text{OOD}}$, its trajectory under the noising process $q(\mathbf{x}_t | \mathbf{x}_0)$ will induce greater errors when estimating the denoising score using the pre-trained model. We propose quantifying these deviations not just at the trajectory level, but also by incorporating structural relevance via pixel-level weighting.

\subsubsection{Quantifying trajectory dynamics}
Let $\mathbf{x}_t$ denote the noisy image at timestep $t$ obtained from the forward diffusion process $q(\mathbf{x}_t | \mathbf{x}_0)$, and let $\boldsymbol{\epsilon}(t)$ denote the true noise. The per-timestep pixelwise error is computed as
\begin{equation}
\text{mse}(\mathbf{x}_0, t) = \left\| \boldsymbol{\epsilon}_\theta(\mathbf{x}_t, t) - \boldsymbol{\epsilon}(t) \right\|_2^2.
\end{equation}
Similarly to previous work, we employ a six-dimensional anomaly score. However, in our case, the calculations were based upon higher-order terms and the temporal behavior of the reconstruction. We briefly recap these concepts and argue their relevance for these settings.
\\
\par
\textbf{Higher order terms:} The first, second and third norm are extracted from the estimated Stein score errors. The different norms can provide more information on the raw magnitude, energy of the squared errors and the severity of outlier errors.
\\
\par
\textbf{Temporal behavior:} The errors only provide information about the current timestep. However, the score errors over time is also a crucial component that can provide a notion of the trajectories. Hence, the time derivative per timestep, together with the first, second and third norm are evaluated.
\\
\par
\textbf{6-dimensional score:}
{DiffPathV2} builds directly upon the original DiffPath~\cite{heng2024out}, which computes a six-dimensional anomaly score from the predicted score $\boldsymbol{\epsilon}_\theta$. Unlike the original formulation, which relies solely on $\boldsymbol{\epsilon}_\theta$, instead we compute the 6D score based on the \textit{ mean squared error} (MSE) between the predicted and ground-truth noise $\boldsymbol{\epsilon}$, offering a more principled signal to assess the accuracy of the score. 
To capture both the error magnitudes and their temporal dynamics, we construct the following six-dimensional score
\begin{equation}
\mathbf{s}_{\text{6D}}(\mathbf{x}_0) =
\left[
\begin{array}{c}
\sum_t \| \text{mse}(\mathbf{x}_0, t) \|_1 \\
\sum_t \| \text{mse}(\mathbf{x}_0, t) \|_2 \\
\sum_t \| \text{mse}(\mathbf{x}_0, t) \|_3 \\
\sum_t \left\| \partial_t \text{mse}(\mathbf{x}_0, t) \right\|_1 \\
\sum_t \left\| \partial_t \text{mse}(\mathbf{x}_0, t) \right\|_2 \\
\sum_t \left\| \partial_t \text{mse}(\mathbf{x}_0, t) \right\|_3
\end{array}
\right].
\end{equation}
The first three terms capture the aggregated $p$-norms of the prediction error across timesteps, while the remaining three quantify the curvature (i.e., rate of change) of the error along the diffusion path. This formulation provides a richer trajectory signature than previous approaches.

\subsubsection{Incorporating structure}

Although the proposed method improves the fidelity of error estimation, it treats all pixels equally when aggregating the MSE statistics. However, not all regions of the image carry equal semantic weight: fine-grained structures and high-frequency textures are often more indicative of semantic content, and are also harder to model during the denoising process. We address this by incorporating a spatial weighting mechanism based on the {Structural Similarity Index Measure (SSIM)}. Specifically, we compute a pixel-wise SSIM map between the input image $\mathbf{x}_0$ and the accumulated predicted scores $\sum_t \boldsymbol{\epsilon}_\theta(\mathbf{x}_t, t)$. The SSIM map acts as a structural relevance filter, or amplifier, localizing regions where the noise estimation might be poor. We then re-weigh the 6D score components by modulating the MSE with the inverted SSIM map as
\begin{equation}
    s(\mathbf{x}_0) = \mathbf{s}_{\text{6D}}(\mathbf{x}_0) \cdot \left( \mathbf{1} - \text{SSIM}\left(\mathbf{x}_0, \textstyle\sum_t \boldsymbol{\epsilon}_\theta(\mathbf{x}_t, t)\right) \right).
\end{equation}
This yields a final anomaly score that emphasizes structurally relevant error regions, thus increasing sensitivity to subtle but semantically meaningful deviations. Importantly, this modulation occurs before global pooling into the 6D summary, ensuring that the influence of salient regions is preserved throughout the scoring process.

{
\centering
\begin{algorithm}[t]
\small
\caption{Inference procedure of \textsc{DiffPathV2}}\label{alg:diffpathv2}
\begin{algorithmic}[1]
\Require Test samples $\mathcal{X}_{\text{test}}$, ID val set $\mathcal{X}_{\text{val}}$, trained DM $ g_\theta$
\Ensure OOD scores for test samples $S_\theta(\mathcal{X}_{\text{test}})$
\vspace{0.8mm}
\Function{ComputeScore}{$\mathbf{x}_0, \boldsymbol{\epsilon}, \boldsymbol{\epsilon}_\theta$}
\LineComment{Inputs are assumed to be normalized}
\vspace{0.8mm}
    \State $\text{MSE}_t \gets \texttt{MSE}(\boldsymbol{\epsilon}, \boldsymbol{\epsilon}_\theta)$ \Comment{shape $(N, T, C, H, W)$}
    \State $\text{SSIM} \gets \texttt{SSIMMap}(\mathbf{x}_0, \sum_t \boldsymbol{\epsilon}_\theta)$ \Comment{shape $(N, C, H, W)$}
    \For{$p \in \{1, 2, 3\}$}
        \State $\mathbf{s}_p \gets \sum_{c,h,w} \left( \sum_t \text{MSE}_t^p \cdot (1-\text{SSIM}) \right)$
        \State $\mathbf{s}_{p+3} \gets \sum_{c,h,w} \left( \sum_t \left( \partial_t \text{MSE}_t^p \right) \cdot (1-\text{SSIM}) \right)$
    \EndFor
    \State \Return $\texttt{stack}([\mathbf{s}_1, \ldots, \mathbf{s}_6])$ \Comment{shape $(N, 6)$}
\EndFunction
\vspace{0.8mm}
\For{$\mathbf{x}_0$ in $\mathcal{X}_{\text{val}}$}
    \State $\{\boldsymbol{\epsilon}_\theta(\mathbf{x}_t, t), \boldsymbol{\epsilon}(t)\} \gets \texttt{DDIMInversion}(\mathbf{x}_0,  g_\theta)$
    \State $\mathbf{s} \gets \texttt{ComputeScore}(\mathbf{x}_0, \boldsymbol{\epsilon}, \boldsymbol{\epsilon}_\theta)$
    \State Append $\mathbf{s}$ to $L_{\text{val}}$
\EndFor
\vspace{0.8mm}
\State Fit GMM: $p_{\text{val}} \gets \texttt{GMM}(L_{\text{val}})$
\vspace{0.8mm}
\For{$\mathbf{x}_0$ in $\mathcal{X}_{\text{test}}$}
    \State Repeat lines 12–13 to compute $\mathbf{s}$ and append to $L_{\text{test}}$
\EndFor
\vspace{0.8mm}
\State \Return $p_{\text{val}}(L_{\text{test}})$
\end{algorithmic}
\end{algorithm}
\par
}

\subsection{Benchmarks}

To evaluate the anomaly detection capabilities of our proposed method, we adopt five publicly available image datasets: CIFAR-10~\cite{krizhevsky2009learning}, CIFAR-100~\cite{krizhevsky2009learning}, SVHN~\cite{netzer2011reading}, CelebA~\cite{liu2015faceattributes}, and Textures~\cite{cimpoi14describing}. Each benchmark is constructed by designating one dataset as in-distribution~(ID) and treating the others as sources of out-of-distribution~(OOD) anomalies. Specifically, we consider CIFAR-10, SVHN, and CelebA as inlier datasets for their respective benchmarks. These combinations are chosen to represent both \emph{near} and \emph{far} semantic shifts. For instance, CIFAR-10 vs. CIFAR-100 represents a near-semantic shift within natural images, while SVHN vs. CelebA exemplifies a far-semantic shift (digits vs. faces). As the primary evaluation metric, we report the Area Under the Receiver Operating Characteristic curve~(AUROC), which measures the ability to distinguish between ID and OOD samples across all thresholds.

We follow standard conventions established in previous work and briefly outline the procedure for completeness. We used the train-validation-test splits provided by each dataset. The images are resized to 64$\times$64 for experiments involving the ImageNet pre-trained model and to 32$\times$32 for other models. During inference, anomaly scores are computed in the test set, while the validation set is used to fit a lightweight GMM, as described in \Cref{alg:diffpathv2}. The GMM enables assignment of log-likelihood scores to test samples to obtain the final anomaly score. The optimal parameters for the GMM are found through a grid search. All experiments are run on an NVIDIA H100 GPU.

\subsection{Baselines}

We compare our proposed method, \textsc{DiffPathV2}, against a diverse set of state-of-the-art (SOTA) approaches that span several modeling paradigms. Baselines can be categorized into two types: foundational models and non-foundational models, requiring dataset-specific training.
\subsubsection{Non-foundational}
We can further categorize the non-foundational approaches in to three distinct baselines: energy-based, flow-based, diffusion-based, and foundation model-based methods.
\\
\par
\textbf{Energy-based Models:} This approach trains a model to assign low energy (or high likelihood) to in-distribution samples. \textit{IGEBM}~\cite{du2019implicit} and \textit{VAEBM}~\cite{xiao2021vaebm} combine energy modeling with generative inference, while \textit{Improved CD}~\cite{du2021improved} stabilizes contrastive divergence training to improve anomaly detection performance.
\\
\par
\textbf{Flow-based Models:} The methods in this group compute exact likelihoods using bijective invertible transformations, also known as Normalizing Flows~\cite{rezende2015variational, dinh2014nice}. \textit{IC}~\cite{Serrà2020Input} calculates the input complexity. \textit{DoS}~\cite{morningstar2021density} uses nonparametric density estimators to measure model statistics. \textit{WAIC}~\cite{choi_waic_2019} leverages model variance across ensemble samples to penalize unreliable likelihood estimates. \textit{TT}~\cite{nalisnick2019detecting} takes the information-theoretic perspective, arguing that the entropy of the samples compared to the generating distribution (i.e., the `typicality') can indicate whether a sample is anomalous. Finally, \textit{LR}~\cite{ren2019likelihood} is a log-likelihood ratio baseline.
\\
\par
\textbf{Diffusion-based Models:} Such methods adapt score-based or DDPMs. \textit{NLL} computes exact log-likelihoods using probability flow ODEs~\cite{song2021scorebased}. \textit{IC (DM)}~\cite{Serrà2020Input} applies input complexity scores to diffusion models. \textit{MSMA}~\cite{mahmood2021multiscale} utilizes multi-scale score matching. \textit{DDPM-OOD}~\cite{graham2023denoising} uses a conditional reconstruction-based approach. \textit{LMD}~\cite{liu2023unsupervised} introduces a masking and inpainting strategy for an improved distance-based method.

\subsubsection{Foundational}
These methods leverage pre-trained generative models without task-specific retraining. \textit{ITD}~\cite{kong2023informationtheoretic} computes the exact NLL in terms of an optimal denoiser as a function of the signal-to-noise ratio. \textit{DiffPath}~\cite{heng2024out} is the baseline method discussed in the theoretical background of this paper. To the best of our knowledge, DiffPath is state-of-the-art on foundational anomaly detection at the time of writing this manuscript.

All baselines are evaluated according to a consistent protocol across benchmarks. For fairness, we reimplement ITD using the same model architecture and pre-processing as DiffPath-based methods.

\section{Experiments}
\label{sec:results}

\subsection{Main results}
The results of our experiments on each dataset are presented in Table~\ref{tab:main}. The quantitative scores evaluate the distinctive power of each model between several datasets. Each value indicates the AUROC and bold numbers indicate the best score per dataset combination. At first glance, it is clear that on average, over all combinations, our proposed improvement methodology performs best. Furthermore, compared to {DiffPath}, \textsc{DiffPathV2} achieves notably higher performance in the near-OOD setting, specifically when distinguishing CIFAR-10 inliers from CIFAR-100 anomalies, where semantic overlap between classes makes detection particularly challenging. Interestingly enough, the runner-up model MSMA, indicated with an underscore, is also a score-based diffusion model. Although MSMA is not foundational, the results still confirm the fact that Stein scores are an appropriate source of information about the datasets. Overall, our results confirm the hypothesis that the trajectory of the diffusion process contains sufficient information to near-perfectly distinguish datasets. Moreover, the proposed method shows that this paradigm translates to even unseen datasets.

\subsection{Ablation on Stein scores}
% To what extent do Stein \textit{errors} and SSIM influence the OOD performance? We establish an ablation experiment where we compare the baseline DiffPath with pure score errors and errors + SSIM. The results are presented in Table~\ref{tab:ablation1}. It can be observed that simply determining the error versus plain Stein scores is not sufficient. Only after the addition of the SSIM do the scores improve. This confirms the hypothesis that the SSIM enhances important areas in the denoising process that provide a notion of the data texture and semantics.

\begin{figure}[!h]
    \centering
    \begin{subfigure}[t]{0.9\linewidth}
        \centering
        \includegraphics[width=0.85\linewidth]{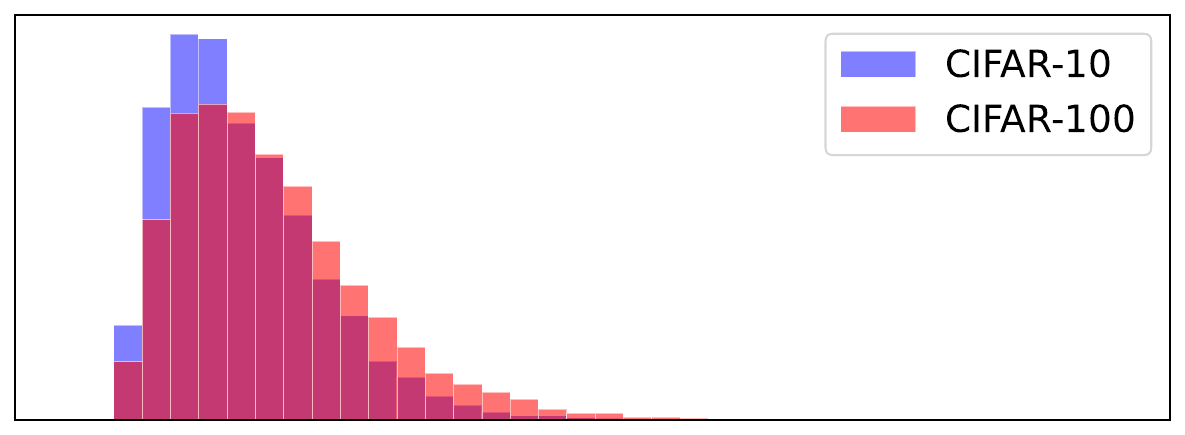}
        \caption{Baseline}
        \label{fig:base}
    \end{subfigure}

    % \vspace{0.5em}

    \begin{subfigure}[t]{0.9\linewidth}
        \centering
        \includegraphics[width=0.85\linewidth]{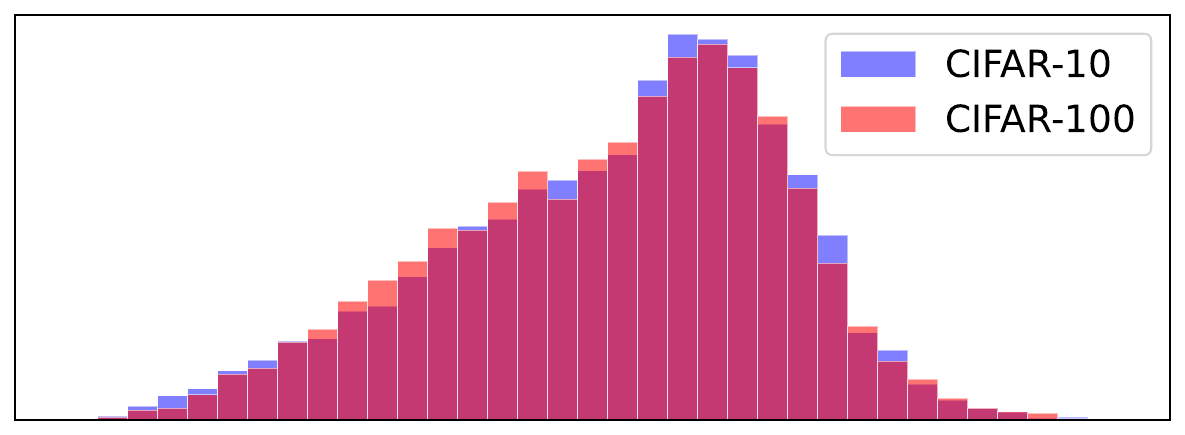}
        \caption{Baseline $\cdot$ (1 - SSIM)}
    \label{fig:ssimbase}
    \end{subfigure}

    % \vspace{0.5em}

    \begin{subfigure}[t]{0.9\linewidth}
        \centering
        \includegraphics[width=0.85\linewidth]{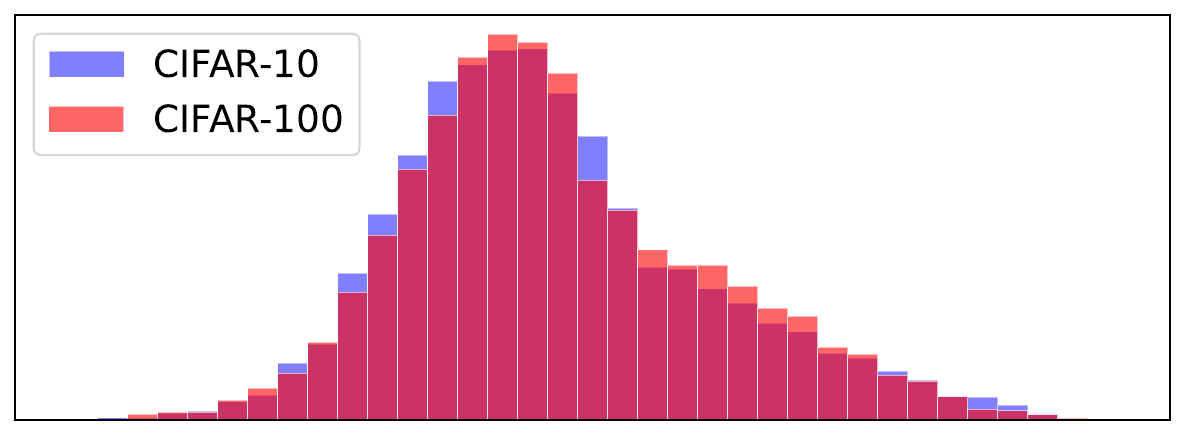}
        \caption{Error}
        \label{fig:error}
    \end{subfigure}

    % \vspace{0.5em}

    \begin{subfigure}[t]{0.9\linewidth}
        \centering
        \includegraphics[width=0.85\linewidth]{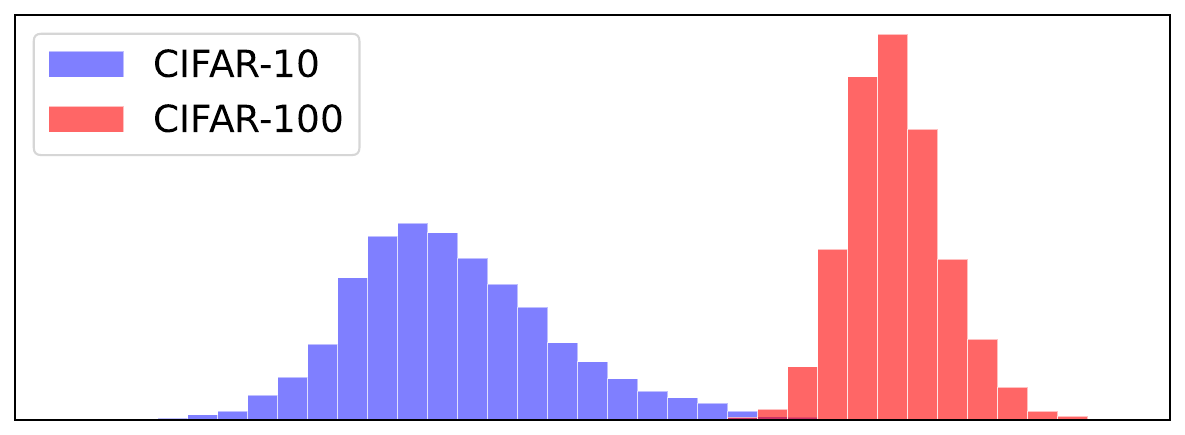}
        \caption{Error $\cdot$ (1 - SSIM)}
    \end{subfigure}

    \caption{Qualitative comparison of Stein score errors and SSIM on Near-OOD (CIFAR10 vs. CIFAR100) performance.}
    \label{fig:ablation_c10}

\end{figure}

To what extent do Stein \textit{errors} and SSIM influence the OOD performance? We conduct an ablation study comparing four variants: the original \textsc{DiffPath} baseline using estimated scores, the same baseline modulated by $(1 - \mathrm{SSIM})$, MSE-based error signals between predicted and true noise, and finally our full method combining error signals with SSIM weighting. Results across all benchmarks are shown in Table~\ref{tab:ablation1}, and the corresponding histograms of anomaly scores are visualized in Figure~\ref{fig:ablation_c10} for the near-OOD setting of CIFAR-10 vs. CIFAR-100.

The baseline without error computation or SSIM already performs competitively on most benchmarks, except under near-semantic shifts as shown in Figure~\ref{fig:base}. However, directly incorporating score errors --- i.e., using the mean squared error between $\boldsymbol{\epsilon}_\theta$ and $\boldsymbol{\epsilon}$ --- generally improves performance, particularly on datasets such as CelebA and SVHN, where semantic anomalies are further from the inlier distribution. Interestingly, modulating the original baseline with the inverse SSIM (\textit{Baseline $\cdot$ (1-SSIM)}) degrades performance, suggesting that SSIM alone does not suffice as a spatial discriminator without complementary error signals. Moreover, it shifts both inlier and OOD distributions toward higher anomaly scores, as seen in Figure~\ref{fig:ssimbase}. In contrast, combining the Stein error with SSIM (\textit{Error $\cdot$ (1-SSIM)})  variant sharply shifts only the OOD distribution to the right while preserving the inlier distribution, resulting in a clear distinction between the two datasets. This method yields the strongest results overall, with an average AUROC of 94.9. These findings confirm the hypothesis that SSIM helps localize regions with meaningful semantic deviations during the denoising process, thereby amplifying anomalies that would otherwise be diluted in global averaging.

Taken together, these results validate that (1)~computing the error between predicted and ground-truth noise estimates provides a more informative signal than raw scores, and (2)~that spatial modulation via SSIM further enhances discriminative power by highlighting perceptually significant regions.

\begin{table*}[thbp]
\caption{Comparison between SOTA methods in semantic AD benchmarks. The Generative Foundational Models do not require the model to retrain on each inlier dataset, contrary to the other methods listed. ($^*$) denotes re-implementation.}
\label{tab:main}
\centering
\resizebox{\textwidth}{!}{%
\addtolength{\tabcolsep}{-0.4em}
\renewcommand{\arraystretch}{1.3} 
\begin{tabular}{@{}cccccccccccccc@{}}
\toprule 

\multicolumn{1}{l}{} &
  \multicolumn{4}{c}{\textbf{C10} {vs.}} &
  \multicolumn{4}{c}{\textbf{SVHN} {vs.}} &
  \multicolumn{4}{c}{\textbf{CelebA} {vs.}} &
  \multicolumn{1}{l}{} \\ \cmidrule(lr){2-5}\cmidrule(lr){6-9}\cmidrule(lr){10-13} 
Method &
  SVHN &
  CelebA &
  C100 &
  Textures &
  C10 &
  CelebA &
  C100 &
  Textures &
  C10 &
  SVHN &
  C100 &
  Textures &
  Average \\ \midrule
% \midrule
\multicolumn{14}{c}{\textit{Energy-based}} \\
\midrule
IGEBM &
  63.0 &
  70.0 &
  50.0 &
  48.0 &
  - &
  - &
  - &
  - &
  - &
  - &
  - &
  - &
  - \\
VAEBM &
  83.0 &
  77.0 &
  62.0 &
  - &
  - &
  - &
  - &
  - &
  - &
  - &
  - &
  - &
  - \\
Improved CD &
  91.0 &
  - &
  83.0 &
  88.0 &
  - &
  - &
  - &
  - &
  - &
  - &
  - &
  - &
  - \\
    \midrule
\multicolumn{14}{c}{\textit{Flow-based}} \\
\midrule
IC &
  95.0 &
  86.3 &
  73.6 &
  - &
  - &
  - &
  - &
  - &
  - &
  - &
  - &
  - &
  - \\
DoS &
  95.5 &
  \underline{99.5} &
  57.1 &
  - &
  96.2 &
  \textbf{100} &
  96.5 &
  - &
  94.9 &
  99.7 &
  95.6 &
  - &
  92.8  \\
WAIC &
  14.3 &
  92.8 &
  53.2 &
  - &
  80.2 &
  99.1 &
  83.1 &
  - &
  50.7 &
  13.9 &
  53.5 &
  - &
  60.1 \\
TT &
  87.0 &
  84.8 &
  54.8 &
  - &
  97.0 &
  \textbf{100} &
  96.5 &
  - &
  63.4 &
  98.2 &
  67.1 &
  - &
  83.2  \\
LR &
  06.4 &
  91.4 &
  52.0 &
  - &
  81.9 &
  91.2 &
  77.9 &
  - &
  32.3 &
  02.8 &
  35.7 &
  - &
  52.4  \\
  \midrule
\multicolumn{14}{c}{\textit{Diffusion-based}} \\
\midrule
  NLL &
  09.1 &
  57.4 &
  52.1 &
  60.9 &
  \underline{99.0} &
  \underline{99.9} &
  \underline{99.2} &
  98.3 &
  81.4 &
  10.5 &
  78.6 &
  80.9 &
  68.9  \\
  IC &
  92.1 &
  51.6 &
  51.9 &
  55.3 &
  08.0 &
  02.8 &
  10.0 &
  17.4 &
  48.5 &
  97.2 &
  51.0 &
  55.9 &
  45.1 \\
MSMA &
  95.7 &
  \textbf{100} &
  61.5 &
  \underline{98.6} &
  97.6 &
   {99.5} &
  98.0 &
  \underline{99.6} &
  91.0 &
  99.6 &
  92.7 &
  \textbf{99.9} &
  \underline{94.5}  \\
DDPM-OOD &
  39.0 &
  65.9 &
  53.6 &
  59.8 &
  95.1 &
  98.6 &
  94.5 &
  91.0 &
  79.5 &
  63.6 &
  77.8 &
  77.3 &
  74.6  \\
LMD &
  \textbf{99.2} &
  55.7 &
  60.4 &
  66.7 &
  91.9 &
  89.0 &
  88.1 &
  91.4 &
  \underline{98.9} &
  \textbf{100} &
  \underline{97.9} &
  \underline{97.2} &
  86.5  \\
\midrule
\multicolumn{14}{c}{\textit{Generative Foundation Models}} \\
\midrule

ITD$^*$ & \underline{99.0} & \textbf{100}  & \textbf{100}  & \textbf{99.0}  & 1.2& 76.8& 57.9 &  58.6   &  91.0 & 62.4 & 52.4 & 58.1  & 76.2  \\ 

DiffPath  & 91.0          & 89.7 & 59.0 & {92.3} & 93.9 & 97.9 & 95.3 & 98.1 & \textbf{99.8} & \textbf{100} & \textbf{99.8} & \textbf{99.9} & 93.1 \\
% DiffPathV2~(Ours) & 93.4 & 96.2 & 52.7 & 91.0 & \textbf{94.4} & 95.8 & 94.6 & 98.8 & \textbf{99.9} & \textbf{100} & \textbf{99.9} & \textbf{99.9} & 93.0 \\
DiffPathV2~(Ours) & 94.9 & \underline{99.5} & \underline{99.4} & 90.5 & \textbf{100} & \textbf{100} & \textbf{100} & \textbf{100} & 91.1 & \textbf{100} & 79.8 & 80.2 & \textbf{94.9} \\
\bottomrule
\end{tabular}}
\end{table*}   

\begin{table*}[!tbp]
\caption{Effect of the Stein score errors and SSIM on AD performance.}
\label{tab:ablation1}
\centering
\resizebox{\textwidth}{!}{%
\addtolength{\tabcolsep}{-0.4em}
\renewcommand{\arraystretch}{1.3} 
\begin{tabular}{@{}cccccccccccccc@{}}
\toprule
\multicolumn{1}{l}{} &
  \multicolumn{4}{c@{}}{\textbf{C10} vs.} &
  \multicolumn{4}{c@{}}{\textbf{SVHN} vs.} &
  \multicolumn{4}{c@{}}{\textbf{CelebA} vs.} &
  \multicolumn{1}{l}{} \\ 
  %[-3.5ex]
  \cmidrule(lr){2-5}\cmidrule(lr){6-9}\cmidrule(lr){10-13} 
Method &
  SVHN &
  CelebA &
  C100 &
  Textures &
  C10 &
  CelebA &
  C100 &
  Textures &
  C10 &
  SVHN &
  C100 &
  Textures &
  Average \\ \midrule 

Baseline  & 91.0          & 89.7 & \underline{59.0} & \textbf{92.3} & 93.9 & {97.9} & {95.3} & {98.1} & \underline{99.8} & \textbf{100} & \underline{99.8} & \textbf{99.9} & \underline{93.1} \\

Baseline $\cdot$ (1-SSIM)  & 92.1 & 77.0 & 58.9 &53.7&\underline{95.7}&\underline{99.2}&\underline{95.8}&91.4&64.4&99.8&68.5&63.6& 80.0\\

Error & 93.4 & \underline{96.2} & 52.7 & \underline{91.0} &  {94.4} & 95.8 & 94.6 & \underline{98.8} & \textbf{99.9} & \textbf{100} & \textbf{99.9} & \textbf{99.9} & 93.0 \\

Error $\cdot$ (1-SSIM)  & \textbf{94.9} & \textbf{99.5} & \textbf{99.4} & 90.5 & \textbf{100} & \textbf{100} & \textbf{100} & \textbf{100} & 91.1 & \textbf{100} & 79.8 & 80.2 & \textbf{94.9} \\
\bottomrule
\end{tabular}}
\end{table*} 

\begin{table*}[thbp]
\caption{Comparison between Generative Foundation Models and the effect of the training dataset.}
\label{tab:ablation}
\centering
\resizebox{\textwidth}{!}{%
\addtolength{\tabcolsep}{-0.4em}
\renewcommand{\arraystretch}{1.3} 
\begin{tabular}{@{}cccccccccccccc@{}}
\toprule
 &
  \multicolumn{4}{c}{\textbf{C10} {vs.}} &
  \multicolumn{4}{c}{\textbf{SVHN} {vs.}} &
  \multicolumn{4}{c}{\textbf{CelebA} {vs.}} & \\ 
  \cmidrule(lr){2-5}\cmidrule(lr){6-9}\cmidrule(lr){10-13} 
Model / $p_{\text{train}}(\mathbf{x})$ &
  SVHN &
  CelebA &
  C100 &
  Textures &
  C10 &
  CelebA &
  C100 &
  Textures &
  C10 &
  SVHN &
  C100 &
  Textures &
  Average \\ \midrule 
ITD / ImageNet & \textbf{99.0} & \textbf{100}  & \textbf{100}  & \textbf{99.0}  & 1.2& 76.8& 57.9 &  58.6   &  91.0 & 62.4 & 52.4 & 58.1  & 76.2  \\ 
% ITD-C10~(Ours) & 98.9 & \textbf{100} & \textbf{100} & 99.1 & 1.8 & 75.9 & 69.3 & 53.8 &91.1& 62.4& 53.2& 58.1& 72.0\\
ITD / CelebA & 1.2 & 0.0 & {0.0} & {0.6} & \underline{98.9} & 20.1 & 27.6 & 45.4 & \textbf{100.0} & {79.9} & {59.7} & {77.3} & 42.6\\
DiffPath / ImageNet  & 85.6          & 50.2 & 58.0 & 84.1 & {94.3} & {96.4} & \underline{95.4} & 96.9 & 80.7 & {98.1} & 84.3 & \underline{96.4} & 85.0 \\

DiffPath / CelebA  & 91.0          & 89.7 & 59.0 & \underline{92.3} & 93.9 & \underline{97.9} & {95.3} & {98.1} & \underline{99.8} & \textbf{100} & {99.8} & \underline{99.9} & \underline{93.1} \\

DiffPathV2 / ImageNet & 08.6 & 63.8 & 00.1 & 96.0 &\textbf{100}& \textbf{100} & \textbf{100} & \textbf{100} & 74.4&\textbf{100} & 68.9&\textbf{100}&76.0\\

DiffPathV2 / CelebA  & 94.9 & \underline{99.5} & \underline{99.4} & 90.5 & \textbf{100} & \textbf{100} & \textbf{100} & \textbf{100} & 91.1 & \textbf{100} & 79.8 & 80.2 & \textbf{94.9} \\
\bottomrule
\end{tabular}}
\end{table*} 

\newpage
\newpage
\subsection{Ablation on Foundational dataset}
Which dataset serves as the best foundational dataset? To assess how the base distribution influences anomaly detection performance, we ablate the effect of the dataset used to pre-train the GFM. Specifically, we compare ImageNet and CelebA as base distributions across multiple GFM. All datasets are resized to a consistent resolution, the higher resolution datasets are downscaled to 32$\times$32, before upscaling all datasets to match the 64$\times$64 ImageNet model using bilinear pixel interpolation.
\\
\par
\textbf{Heterogeneity is not always better:}
The field often enjoys prototypical large-scale and diverse datasets such as ImageNet, which offers broad generalization. However, we find that training on CelebA consistently yields better average anomaly detection scores for Diffpath(V2), as reported in \Cref{tab:ablation}. Our findings suggest that increasing the heterogeneity of the foundational distribution does not always lead to better performance, particularly when the anomaly score is tied to fine-grained deviations along the generative trajectory. These findings validate previous experiments in the literature~\cite{heng2024out}, which evidently translate to the proposed method. 
% 
% \paragraph{...}
CelebA may offer a form of structural consistency that improves the sensitivity of the model to perturbations in the diffusion process, despite its narrower semantic scope. Our method explicitly measures deviations in the trajectory curvature and amplifies this via the SSIM. Therefore, such semantic coherence appears to reinforce the scoring mechanism. This indicates that certain datasets, although semantically narrower, might serve as a better base distribution. 
\\
\par
\textbf{Foundational datasets are objective-dependent:}
In contrast, the Information-Theoretic Diffusion~(ITD) model performs best with the ImageNet base and shows limited generalization when trained on CelebA. This is likely due to its direct optimization of the exact log-likelihood $\log p(\mathbf{x})$ as a function of the signal-to-noise ratio (SNR), which is tightly coupled to the distributional complexity of the training domain. While this formulation achieves SOTA results for density estimation~\cite{kong2023informationtheoretic}, it may be less robust to anomaly detection across multiple benchmarks.

These findings highlight that the choice of pretraining data not only defines the generative capacity of the model but also determines how well samples from different distributions can be detected in it's diffusion trajectory. Rather than simply suggesting that narrower datasets are better, we emphasize that there is a complex interplay between the diversity of the base distribution and the nature of the anomaly detection objective.

\section{Conclusion}
\label{sec:conclusion}

In this work, we demonstrate that score-based models can serve as foundational generative models for semantic anomaly detection across datasets never seen during training. Specifically, we show that the denoising trajectory of Denoising Diffusion Models (DDMs) encodes rich information about both texture and semantics. We extract and analyze statistical signals from Stein score errors, amplified using the Structural Similarity Index Metric (SSIM), and leverage these to detect anomalous samples and improve over the state-of-the-art. The implemented methods are truly foundational, as they are trained on solely a single dataset and only utilizes an in-distribution validation set for determining an appropriate error threshold. Contrary to the common reliance on ImageNet, we find that CelebA provides a surprisingly effective base distribution for a variety of anomaly detection baselines. This work underscores the feasibility and promise of using generative foundation models for anomaly detection. While near-perfect scores are achieved on some datasets, significant room for improvement remains on others. We encourage further exploration of anomaly detection using (larger) generative foundation models, particularly by leveraging the denoising trajectories of score-based models.

\vfill

{
    \small
    \bibliographystyle{ieeenat_fullname}
    \bibliography{ref}
}

\end{document}